\documentclass{article}

\usepackage{arxiv}
\usepackage[utf8]{inputenc} % allow utf-8 input
\usepackage[T1]{fontenc}    % use 8-bit T1 fonts
\usepackage{hyperref}       % hyperlinks
\usepackage{url}            % simple URL typesetting
\usepackage{booktabs}       % professional-quality tables
\usepackage{amsfonts}       % blackboard math symbols
\usepackage{nicefrac}       % compact symbols for 1/2, etc.
\usepackage{microtype}      % microtypography
\usepackage{xcolor}         % colors
\usepackage{enumitem}
\usepackage{bm}
\usepackage{amsmath}
\usepackage{amssymb}
\usepackage{amsthm}
\usepackage{amsmath, amssymb, amsthm}
\usepackage{graphicx}
\usepackage{epstopdf}
\usepackage{svg}
\usepackage{subcaption}

\usepackage{algorithmic}
\usepackage{algorithm}
\newtheorem{theorem}{Theorem}
\title{Spatio-spectral graph neural operator for solving computational mechanics problems on irregular domain and unstructured grid}

\author{
  Subhankar Sarkar \\
  Yardi School of Artificial Intelligence\\
  Indian Institute of Technology Delhi\\
 Hauz Khas, New Delhi 110016 \\
  \texttt{subhankar.sarkar@scai.iitd.ac.in} \\
  \And
  Souvik Chakraborty \\
  Department of Applied Mechanics\\
  Yardi School of Artificial Intelligence\\
  Indian Institute of Technology Delhi\\
 Hauz Khas, New Delhi 110016 \\
  \texttt{souvik@am.iitd.ac.in} \\
}

\begin{document}

\maketitle

\begin{abstract}
Scientific machine learning has seen significant progress with the emergence of operator learning. However, existing methods encounter difficulties when applied to problems on unstructured grids and irregular domains. Spatial graph neural networks utilize local convolution in a neighborhood to potentially address these challenges, yet they often suffer from issues such as over-smoothing and over-squashing in deep architectures. Conversely, spectral graph neural networks leverage global convolution to capture extensive features and long-range dependencies in domain graphs, albeit at a high computational cost due to Eigenvalue decomposition. 
In this paper, we introduce a novel approach, referred to as Spatio-Spectral Graph Neural Operator (Sp$^2$GNO) that integrates spatial and spectral GNNs effectively. This framework mitigates the limitations of individual methods and enables the learning of solution operators across arbitrary geometries, thus catering to a wide range of real-world problems. Sp$^2$GNO demonstrates exceptional performance in solving both time-dependent and time-independent partial differential equations on regular and irregular domains. Our approach is validated through comprehensive benchmarks and practical applications drawn from computational mechanics and scientific computing literature.
\end{abstract}

% \begin{keyword}
% % % Keywords for the paper. Write "\sep" between two keywords. Example given below.
% Operator Learning \sep Graph Neural Network \sep Spectral Graph Neural Network \sep Neural Operator \sep  Computational
% Mechanics
% \end{keyword}

\section{Introduction}
Modeling and analysis of a physical system often require solving a system of Partial Differential Equations (PDEs) subjected to certain initial and/or boundary conditions. Unfortunately, most of the PDEs are not solvable analytically; thus, numerical solvers like Finite Difference \cite{finite_diffrence}, Finite Element \cite{finite_element}, and Finite Volume \cite{EYMARD2000713}  methods are used in practice. These solvers are computationally expensive; thus, they take a considerable amount of time, even on modern computers. Apart from that, these solvers are discretization-dependent and hence, for any changes in domain discretization and/or parameter values, the solver must recalculate the solution from scratch. This makes the overall process computationally tedious and time-consuming, which limits their applicability in engineering design. Recent advancements in deep learning using deep neural networks (DNNs) have paved the way toward data-driven alternatives to classical numerical solvers. DNNs are \textit{univeral approximators} and capable of approximating any function. PDE solving process using DNNs can be data-driven \cite{raissi2018deep, Parashar2020, Navaneeth2022, Rudy2017} or physics informed \cite{RAISSI2019686, GOSWAMI2020102447, SAMANIEGO2020112790, CHAKRABORTY2021109942}.
Physics-informed learning attempts to exploit known physics in form of PDE within the loss function.
However, in most real-world scenarios, the exact governing law of physics is not known beforehand, which can potentially limit the applicability of physics-informed algorithms. In these scenarios, the data-driven approach becomes the only alternative. Unlike their physics-informed counterpart, data-driven algorithms cannot learn from the physical law and, thus, don't generalize well beyond the training region. Despite poor generalizability beyond the training region, data-driven algorithms have much broader applicability, specifically in scenarios where data is available beforehand, but the exact physics is unknown \cite{deep_learning_pde, dl_pde, dl_pde2, dl_pde3}. However, conventional DL-based data-driven approaches \cite{dl_pde, dl_pde2, dl_pde3} for computational mechanics are discretization-dependent and

The first operator learning algorithm was proposed in \cite{ChenandChen}. This was later extended to deep networks, and the resulting architecture was referred to as Deep Operator Network (DeepONet) \cite{Deeponet}. DeepONet leverages the universal approximation theorem for operators \cite{ChenandChen} by using the branch and trunk networks. The branch network handles the input, and the trunk network handles the sensor locations. Improvements to the vanilla DeepONet include proper orthogonal decomposition-based DeepONet \cite{Eivazi2024NonlinearMR}, and U-DeepONet \cite{Diab2023UDeepONetUE}. Physics-informed DeepONet \cite{Goswami2022PhysicsInformedDN} can also be found in the literature. DeepONets and its variants have been used for solving a wide array of engineering problems including fracture mechanics \cite{GOSWAMI2022114587}, bubble dynamics \cite{lin2021seamless}, and reliability analysis \cite{Garg2022AssessmentOD}.

Almost parallel to DeepONet, the kernel-based neural operator was proposed in \cite{Li2020NeuralOG}. The idea here is to utilize Green's function-based formulation to develop the operator learning algorithm. Owing to the inspiration from Green's function, this class of operator learning algorithms are particularly suitable for systems governed by partial differential equations. However, exactly evaluating the integration present in the kernel is challenging. The original work approximated the integration by aggregating neighborhood information using a graph neural network and was referred to as the Graph Neural Operator (GNO). Unfortunately, GNO scales poorly with an increase in nodes and has a very high memory requirement. Further, being a spatial graph neural network, it also suffers from over-smoothing \cite{Rusch2023ASO}. To address this apparent challenge, spectral approaches were adopted; the idea here is to solve the kernel integration in the spectral domain. Fourier Neural Operator (FNO) \cite{li2020fourier} and Wavelet Neural Operator (WNO)  are perhaps the most popular spectral neural operators currently existing in the literature. While FNO solves the kernel integration by projecting it onto the Fourier basis, WNO solves the same by projecting it onto the wavelet basis. Variants of FNO such as factorized FNO \cite{Tran2021FactorizedFN}, U-FNO \cite{Wen2021UFNOA}, and variants of WNO such as Waveformer \cite{Navaneeth2023WaveformerFM}, generative WNO \cite{Soin2024GenerativeFI, rani2024generative}, and physics-informed WNO \cite{N2023PhysicsIW} can also be found in the literature. All the spectral-based approaches discussed above are highly accurate; however, these are naturally designed to handle regular domain and structured grids. This limits their applicability for problems defined on irregular domains and unstructured grids.

To address the above challenge, geometry-aware FNO (geo-FNO) \cite{Li2022FourierNO} 
was proposed. In geo-FNO, a separate neural network is used, 
which learns to deform the input (physical) domain into a latent space with a uniform grid. The FNO is then employed in the latent space. Although geo-FNO yields encouraging results, it is limited in scope as, for complex geometries, diffeomorphism from the physical space to the uniform computational space does not necessarily exist. In response to this challenge, this paper aims to develop an operator-learning algorithm that can exploit the advantages of spectral neural operators, scale efficiently, and seamlessly handle irregular domains and unstructured grids. The primary bottleneck towards achieving this stems from the inherent limitation associated with Graph Neural Networks (GNN) \cite{Hamilton2017InductiveRL}. Specifically, deep GNNs suffer from the over-smoothing problem, severely degrading their performance. Besides, spatial GNNs can only capture the local short-range dependencies, severely limiting their learning ability. Spectral GNNs \cite{Bruna2013SpectralNA, Liao2019LanczosNetMD, Defferrard2016ConvolutionalNN} overcome these by using spectral graph convolution, which learns by projecting the spatial features to the space of eigenvectors of the graph Laplacian. The spectral graph convolution is a global convolution that can capture the long-range dependencies within the graph owing to the superior expressivity of spectral GNNs.  However, spectral GNNs are computationally expensive because of the prohibitive $O(N^3)$ cost associated with the full eigenvalue decomposition of the graph Laplacian.

In this paper, we propose the Spatio-Spectral Graph Neural Operator (Sp$^2$GNO) that exploits both spatial GNN and spectral GNN to develop an effective operator learning algorithm that encourages collaboration between spatial and spectral GNN so as to overcome individual limitations. The key features of Sp$^2$GNO are highlighted below:

\begin{enumerate}

    \item \textbf{Effective collaborative learning:} Sp$^2$GNO parallelly learns through both spatial and spectral GNNs to capture both long-range and local dependencies and collaborate in a manner that prevents the over-smoothing issue. Sp$^2$GNO treats the domain as a graph, which makes it applicable in any real-world problem with an unstructured grid and irregular domain.
    
    \item \textbf{Expressive and scalable:} To make the network scalable, instead of $O(N^3)$ full eigenvalue decomposition, we use first $m$ eigenvectors as bases, which has $O(mE)$ complexity, where $E$ is the number of edges in the domain graph. Instead of learning a diagonal filter, we learn a 3D kernel in the spectral domain for better expressivity.
    
    \item \textbf{Gating mechanism with Lisfshitz embedding :} We incorporate a gating mechanism in spatial GNN, which takes Lipshitz embedding as an input to the edge-weight gate to make the network position-aware and efficiently learn edge weights to learn the optimal graph.
\end{enumerate}

The remainder of the paper is organized as follows. In Section \ref{sec:Background}, the background and problem setup are discussed. Section \ref{sec:Sp2GNO} discusses the proposed Sp$^2$GNO architecture. Numerical results are presented in Section \ref{sec:Experiments} to illustrate the performance of the proposed approach. Finally, Section \ref{sec:concl} provides the concluding remarks.

\section{Problem Setup and Background}
\label{sec:Background}

\subsection{Problem Setup}

Solving a PDE by learning an operator involves learning the mapping  $\mathcal{M} : \mathcal{A} \to \mathcal{U} $ between two infinite dimensional Banach spaces $\mathcal{A}$ and $\mathcal{U}$ through a deep architecture $F_{\bm {\theta}}$. Here, $\mathcal{A}$ is the space for the input functions, and $\mathcal{U}$ is the space for the output functions. We work with the observed samples of input and output functions $(a^{(i)}(\bm x), u^{(i)}(\bm x))$, which are functions of the coordinates $\bm x$. With slight abuse of notation, we here onwards denote $a(\bm x )$ as $a$ and $u(\bm x ) $ as $u$. The domain $\mathcal{D} \subset \mathbb{R}^d$ is a bounded open set of dimension $d$, where both the input function $a \in \mathcal{A} \subset L^2(D;\mathbb{R}^{d_{x}})$ and output function $u \in \mathcal{U} \subset L^2(D;\mathbb{R}^{d_{y}})$ is defined, with $\mathbb{R}^{d_{x}}$ and $\mathbb{R}^{d_{y}}$ indicating the range of $a \in \mathcal{A}$ and $u \in \mathcal{U}$ respectively. In practice, the domain is discretized in $N$ grid points, which makes $\mathcal{D}$ a finite set containing $N$ grid points. For each grid point $p \in \mathcal{D}$, both the input $a$ and output $u$ have their value in $\mathbb{R}^{d_{x}}$ and $\mathbb{R}^{d_{y}}$ respectively, i.e $a(p) \in \mathbb{R}^{d_{x}}$ and $u(p) \in \mathbb{R}^{d_{y}}$.
To find the optimal set of model parameters $\bm \theta$, we minimize the cost function
\begin{equation}
\bm \theta^* = \arg \min_{\bm \theta} \sum_{i=1}^{N_{train}} C(F_{\bm \theta}(\bm a^{(i)}), \mathcal{M}(\bm a^{(i)})),
\end{equation}

where $C$ is the cost function, which measures the dissimilarity between the ground truth $\mathcal{M}(\bm a^{(i)})$ and the prediction $F_{\bm \theta}(\bm a^{(i)})$.
Thus, during training, 
the objective is to find optimal network parameters, $\bm \theta^*$, such that $\mathcal{M} \approx F_{\bm \theta^*}$.
We train the model $F_{\bm \theta}$ using observed samples $\{(\bm a^{(i)}, \bm p^{(i)}), \bm u^{(i)})\}_{i=1}^{N_{train}}$, where $N_{train}$ is the number of data points in the training set, and $\bm p^{(i)})$ is the sensor location/grid points. Note that each sample $(\bm a^{(i)}, \bm u^{(i)})$ consist of values of $a$ and $u$ at all $N$ grid points $p \in \mathcal{D}$, i.e., $\bm a^{(i)} = \{a^{(i)}(p)\}_{p=1}^N $ and $\bm u^{(i)} = \{ u^{(i)}(p)\}_{p=1}^N$. 
This discretization makes each input and output sample a matrix, i.e., $\bm a^{(i)} \in \mathbb{R}^{N \times d_x}$ and $u^{(i)} \in \mathbb{R}^{N \times d_y}$.

With this setup, the objective of this paper is to develop an architecture $F_{\bm \theta}$ that can handle $\bm a ^{(i)}$ and $\bm u ^{(i)}$ defined on unstructured grid points $\bm p$ and irregular domain $\mathcal D$, in a seamless manner.

\subsection{Neural Operator}
Let's take an example of a generalized PDE, where $\mathcal{T}$ is any differential operator.
\begin{equation}
\label{eq:generalised_pde}
\begin{aligned}
    \mathcal{T} u(x) &= f(x), \quad x \in D, \\
    u(x) &= 0, \quad x \in \partial D.
\end{aligned}
\end{equation}
The bounded open set $D \subset \mathbb{R}^d$ in Eq. \eqref{eq:generalised_pde} is the domain, and $f$ is the forcing function (or the source term). For example, if $\mathcal{T} = -\nabla \cdot (a \nabla \cdot ) $, then our generalized PDE in Eq. \eqref{eq:generalised_pde} will be reduced to the elliptic PDE $ -\nabla \cdot( a \nabla)u(x) = f(x)$.
In case the operator $\mathcal T$ is a linear operator, the solution of Eq. \eqref{eq:generalised_pde} is given by the convolution integral with Green's function $G\left(\cdot;\cdot\right)$,
\begin{equation}
\label{eq:integral_equation}
u(x) = \int_D G(x, z) f(z) \, dz,
\end{equation}
where, $G(x,z)$  satisfies, $\mathcal{T} G(x,z) = \delta_x$. Here, $\delta_x$ is the Dirac delta function centered at $x$. In  kernel-based neural operators\cite{Li2020NeuralOG}, the basic idea is to exploit Greeen's function-based formulation in Eq. \eqref{eq:integral_equation} in conjunction with nonlinearity introduced through activation operator in an iterative fashion, 
% first proposed to approximate this convolution by local spatial convolution in the domain graph iteratively through T convolutional layers. That is, for $t = 0,..., T-1$
\begin{equation}
\label{eq:vt_equation}
v_{t+1}(x) = \sigma \left( W v_t(x) +  \int_D \kappa_\phi(x, z, a(x), a(z)) v_t(z)dz \right),
\end{equation}
where $Wv_t(x)$ is the residual connection through a linear transform added in every layer and $W\in \mathbb{R}^{d \times d}$ is the weight of the linear transform. The spatial kernel $K_\phi$ is similar to Green's function and is to be learned based on training data. The features $v_0(x)$ are constructed by an MLP $P$ from initial features $\{a(p), p\}$. The final output is obtained by passing the output of the $T$th convolutional layer through an MLP $Q$. Therefore, the overall forward pass for any kernel-based neural operator can be represented as,
\begin{equation}
\label{eq:vt_equation2}
\begin{aligned}
v_0(x) &= P(\{a, x\}) \\
v_{t+1}(x) &= \sigma \left( W v_t(x) +  \int_D \kappa_\phi(x, z, a(x), a(z)) v_t(z)dz \right) \\
u(x) &= Q(v_T(x))
\end{aligned}
\end{equation}
This equation serves as a backbone of kernel-based spectral operator learning algorithms, including FNO \cite{li2020fourier} and WNO \cite{TRIPURA2023115783}.

\subsection{Spectral Graph theory and Spectral Graph Convolution}\label{subsec:sgnn}
We define a graph as $G = (\mathbb V, \mathbb X, \mathbb E)$, where $\mathbb V$ is the set of nodes, $\mathbb E$ is the set of edges, and $\mathbb X \in \mathbb R^{N \times d}$ is the node feature matrix containing $d$ dimensional features per node in its rows. The connectivity of the graph is represented by its weighted adjacency matrix $\mathbf A$, where 
\begin{equation}
    A_{ij} = w_{ij}.
\end{equation}
$w_{ij}$ is the weight of the connection between the node $i$ and node $j$. If the adjacency matrix is unweighted, then 
\begin{equation}
   w_{ij} = \left\{ \begin{array}{cl}
        1 & \text{ if there is an edge between node } i \text{ and } j\\
         0 & \text{elsewhere}
    \end{array} \right.
\end{equation}
Symmetrical degree normalized graph Laplacian is defined as 
\begin{equation}\label{eq:laplace}
    \Tilde{\mathbf L} = \mathbf{I} -\mathbf D^{-1/2}\mathbf A \mathbf D^{1/2},
\end{equation}
where $\mathbf D$ is the diagonal degree matrix of the graph obtained by summing the adjacency matrix row-wise, i.e, 
\begin{equation}
    d_{i}= \sum_j \mathbf{A}_{ij}.
\end{equation}
Finally, the spectral decomposition of normalized Laplacian is defined as
\begin{equation}\label{eq:eigen_decomp}
    \Tilde{\mathbf L} =  \mathbf Q \bm \Lambda \mathbf Q^T,
\end{equation}
where $\bm \Lambda \in \mathbb R^{N\times N}$ is a diagonal matrix containing all eigenvalues of the graph Laplacian in its principle diagonal. $\mathbf Q$ is the matrix containing the eigenvectors of graph Laplacian. Similar to the sine and cosine bases of the DFT, the eigenvectors of $\mathbf Q \in \mathbb R^{N\times N}$ serve as bases of Graph Fourier Transform (GFT). Building on this, the Graph Fourier Transform (GFT) and Inverse Graph Fourier transform (IGFT) of the node features are defined as,
\begin{equation}\label{eq:GFT}
\begin{aligned}
\hat{\mathbb X} &= \mathbf Q^T \mathbb X \quad &\text{Graph Fourier Transform} \\
\mathbb X &= \mathbf Q\hat{\mathbb X} \quad &\text{Inverse Graph Fourier Transform} \\
\end{aligned}
\end{equation}
Using Eq. \eqref{eq:GFT}, the spectral graph convolution of the node features with respect to kernel $\mathbf K$ is defined as,
\begin{equation}
\bm y = \mathbf Q(\mathbf Q^T \mathbf K \odot \mathbf Q^T \mathbb X )
\end{equation}
Note that both kernel and node features are first transformed via GFT and then multiplied element-wise, followed by an IGFT. Now, this kernel $\mathbf Q^T \mathbf K$ can be directly parametrized in the spectral domain by a function $g_{\bm \theta}$,
\begin{equation}\label{eq:spectral_conv}
\bm y = \mathbf Q \mathbf g_{\bm \theta} \mathbf Q^T \mathbb X,
\end{equation}
where, $g_{\theta}$ is a diagonal filter learned from data.
This is the spectral convolution used in spectral GNNs \cite{Bruna2013SpectralNA, Defferrard2016ConvolutionalNN}.  Unlike $\mathbf A$ in the case of spatial convolution, the matrix $\mathbf Q$ is not sparse. Thus, the multiplication with $\mathbf Q^T$ and $\mathbf Q$ with $\mathbb X$ aggregates node features from all nodes to construct the feature for a particular node after convolution. Thus, spectral graph convolution is a global convolution and captures the log-range dependencies owing to the high expressiveness of the spectral GNNs. But, the Cubic $O(N^3)$ complexity of eigenvalue decomposition makes spectral GNNs scale poorly to large graphs.

\subsection{Spatial Graph Convolution}
In spatial graph convolution, the node features of the particular node are constructed by aggregating the node features of its direct neighbors or K-hop neighbors. We can define a general spatial graph convolution mathematically as,
\begin{equation}\label{eq:spatial_conv}
\bm y = f( \mathbf A) \mathbb X
\end{equation}
where $ f(\mathbf A)$ is a matrix-valued function of the adjacency matrix, $\mathbf A$ of the graph, $\mathbb X$ is the node features as defined earlier. As the adjacency matrix $\mathbf A$ is sparse, multiplication $f(\mathbf A)$  with $\mathbb X$ only aggregates node features of its direct neighbors. Thus, unlike its spectral counterpart, spatial convolution is a local convolution. Different choices of the function $f(\mathbf A)$ results in several different spatial GNN architectures. The most popular lightweight spatial GNN architecture is the Graph Convolutional Network (GCN) \cite{Kipf2016SemiSupervisedCW}. Here, the function $f(\mathbf A)$ is chosen to be 
\begin{equation}\label{eq:gcnn}
     f(\mathbf A) = {(\mathbf D+ \mathbf I)}^{-1/2}{(\mathbf A + \mathbf I)}(\mathbf D + \mathbf I)^{-1/2},
\end{equation}
where $\mathbf D = \text{diag}(d_i)$ is the diagonal degree matrix defined in Eq. \eqref{eq:laplace} and $\mathbf I$ represents an identity matrix.  The choice of $ f\left(\mathbf A\right)$ in Eq. \eqref{eq:gcnn} results in the spatial graph convolution,
\begin{equation}
   \bm y = {(\mathbf D+ \mathbf I)}^{-1/2}{(\mathbf A+\mathbf I)}(\mathbf D + \mathbf I)^{-1/2}\mathbb X \mathbf W
\end{equation}
where $\bm y$ is the output of the GCN layer, and $\mathbf W$ is the weight matrix associated with that layer. The spatial convolution operation in GCN aggregates the node features from the local 1-hop neighborhood for each node $v_i$. After applying a nonlinearity, the spatial graph convolution is  mathematically represented as,
\begin{equation}
    \bm y_i = \sigma \left( \sum_{j \in \mathcal{N}(i) \cup \{i\}} \frac{1}{\sqrt{d_{i} d_{j}}} \mathbb X_j \mathbf W \right)
\end{equation}
After convolution, the node feature $\bm y_i$ of a node $v_i$ is constructed aggregating the node features $\mathbb X_j$ of all node $v_j \in \mathcal{N}(i) \cup \{v_i\}$, including itself. Here, $\mathcal{N}(i)$ denotes the set containing all the direct 1-hop neighbor nodes of $v_i$. Here, $d_{i}$ and $d_{j}$ are $i$th and $j$th elements along the principle of the diagonal of the matrix $\mathbf D$. 

Stacking too many spatial graph convolutional layers makes the node features of a local neighborhood indistinguishable, leading to information loss and performance degradation. 

\section[Spatial-Spectral Graph Neural Operator (Sp2GNO)]{Spatial-Spectral Graph Neural Operator (Sp$^2$GNO)}
\label{sec:Sp2GNO}
In this section, we propose a novel frequency-based operator learning algorithm that can handle unstructured grids and irregular domains in a seamless manner. Given the fact that GNN, by design, can handle unstructured data in terms of point cloud, the operator learning algorithm proposed in this section attempts to exploit the same. In particular, the high-level idea is to blend spatial and spectral GNNs within the proposed operator learning so that the two can complement each other. The resulting architecture is referred to as spatio-spectral graph neural operator (Sp$^2$GNO).

\subsection[Overall architecture of Sp2GNO]{Overall architecture of Sp$^2$GNO}
In Sp$^2$GNO, we follow the popular architecture of spectral-based operators \cite{li2020fourier, TRIPURA2023115783}, which first encodes the node features and then processes them through a number of spectral-based convolutional layers before passing them through the decoder to produce the final output. Spectral-based operators, by design, can not handle irregular domains. To handle the irregularity in the domain structure, instead of treating the domain as a set of regularly discretized grid points, we first construct a k-nearest neighbor graph connecting each grid point in the domain to its closest k neighbors. The initial input node features $\{\bm a , \bm x\}$ of the graph are constructed by concatenating both the input field $a$ and coordinates $x$. Node features of the input domain-graph are first encoded by using a shallow network $P : \mathbb{R}^{d_{init}} \to \mathbb{R}^d $, where $d_{init}$ is the initial dimension of the node features and $d$ is the hidden dimension of the encoded node-features. The encoded node features $v_0(x)$ are then processed by $L$ number of Sp$^2$GNO blocks, where each block approximates the kernel integration in Eq. \eqref{eq:vt_equation}. The architecture of Sp$^2$GNO block is the primary contribution of the this work.
The Sp$^2$GNO block is designed to encourage collaboration between spatial and spectral GNNs. This is achieved by first processing node features, in parallel, through both the spatial and spectral GNNs and then passing the concatenated outputs from the two GNNs through a linear layer.
Details on the Sp$^2$GNO block are provided in Section \ref{subsec:sp2gno}.
Finally, processed node features $v_L(x)$ are decoded by using another network $Q : \mathbb{R}^d \to \mathbb{R}^{d_{u}} $ to obtain the final output $u$. The overall operation in Sp$^2$GNO is given by,
\begin{equation}
\bm u =  Q\circ S_{L}\circ\cdot\cdot\cdot S_{1}\circ P(\{ \bm a, \bm x\})
\end{equation}
where, $S_{1}$ to $S_{L}$ are the Sp$^2$GNO blocks. $\bm u$, $\bm a$ and  $\bm x$ are the output field, input field, and the coordinates of the domain vertices, respectively.

\begin{figure}[htbp]
    \centering
    {\includegraphics[width=\textwidth]{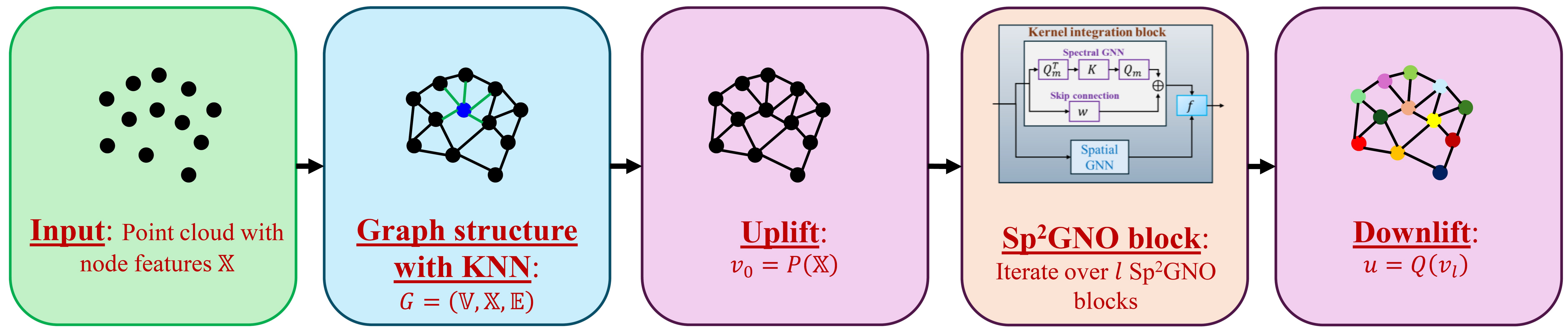}}
    \caption{The overall working principal of the proposed Spatio-Spectral Graph Neural Operator. It involves graph structure generation from point clouds, uplifting the feature, iterating through Sp$^2$GNO block and then downlifting.}
    \label{fig:sp2gno_comparison}
\end{figure}

\subsection[Sp2GNO Block]{Sp$^2$GNO Block}\label{subsec:sp2gno}
\begin{figure}[htbp]
    \centering
    \rotatebox{270}{\includegraphics[width=0.32\textwidth]{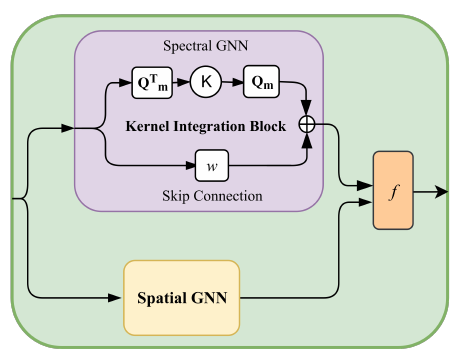}}
    \caption{Neural architecture of a Sp$^2$GNO block. It exploits spectral and spatial graph neural network to formulate the kernel integration operator. The overall architecture involves multiple such blocks.}
    \label{fig:sp2gno_block}
\end{figure}

Sp$^2$GNO block approximates the kernel integration in Eq. \eqref{eq:vt_equation} via both the local and global convolution. We employ a novel spectral GNN based on truncated GFT to perform the global convolution, capturing the long-range information within the domain graph. Parallely, a spatial GNN performs the local convolution, which captures the short-range information while learning the optimal domain graph structure through the gating mechanism over the Lipschitz embeddings. Finally, we concatenate the outputs of both the GNNs and pass the concatenated features through another MLP to get the final output of the Sp$^2$GNO block. The MLP decides the optimal proportion of both the GNNs' contribution in the final output of each Sp$^2$GNO block, facilitating the collaboration between two GNNs. Through this collaboration, the final output of each block will have contributions from both the GNNs, enriching it with both long and short-range information to approximate the kernel integration effectively. This collaboration also mitigates the over-aggregation of the information in each node from its neighborhood, preventing over-smoothing issues. A schematic representation of the Sp$^2$GNO block is shown in Fig. \ref{fig:sp2gno_block}. Further details on the components within the Sp$^2$GNO block is discussed in subsequent sections.

\subsubsection{Global Spectral Graph Convolution based on truncated GFT}
One key player within the Sp$^2$GNO block is the spectral graph convolution discussed in Section \ref{subsec:sgnn}. However, the primary bottleneck with vanilla spectral graph convolution resides in its computational cost. Therefore, within the proposed Sp$^2$GNO block, we have employed a novel spectral GNN based on truncated GFT to approximate the kernel integration in Eq. \eqref{eq:vt_equation}.

\noindent \textbf{Fast and Scalable Spectral Convolution:} Revisiting Eq. \eqref{eq:spectral_conv}, the matrix $\mathbf Q \in \mathbb{R}^{N \times N}$ is the Eigenvector matrix and serves as the basis of the spectral Graph Fourier Transform (GFT) defined in Eq. \eqref{eq:GFT}. However, the computational complexity associated with computing $\mathbf Q$ using Eq. \eqref{eq:eigen_decomp} is $O(N^3)$. This is a major roadblock that prohibits scalability to large graph. In practice, it is observed that the signals that we encounter in physical systems contain the first few lowest frequencies. Accordingly, we exploit a truncated version of GFT for fast and scalable computation that uses the first $m$ eigenvectors as bases of GFT and learn the spectral coefficients by using a learnable kernel $\mathbf K$. This approach helps in bypassing the need for computing the full matrix $\mathbf Q \in \mathbb{R}^{N \times N}$ via full Eigen-decomposition; instead, we partially compute only the first $m$ eigenvectors using \textit{Locally Optimal Block Preconditioned Conjugate Gradient} (LOBPCG) algorithm, resulting in $\mathbf{Q_m} \in \mathbb{R}^{N \times m}$, and the associated computational cost is $O(mE)$. Therefore, for a graph constructed by the k-nearest neighbor (kNN) method with $k$ neighbors, the computational cost of this approach is  $O(mkN)$. Assuming, $m, k<<N$, we have $O(mkN) \approx O(N)$. Further, as the matrix size reduces, the cost of matrix multiplications in GFT is also reduced to  $O(mdN)$ from $O(N^2d)$. Again, considering $m, d<<N$, $O(mdN) \approx O(N)$. Therefore, using truncated GFT gives Sp$^2$GNO better scalability, while preserving the expressivity.

\noindent \textbf{Enriching the Parameter Space:} In the vanilla spectral graph convolution defined in Eq. \eqref{eq:spectral_conv}, the spectral kernel $\bm g_{\bm \theta}$ is a diagonal weight matrix, which ensures learning only one parameter per spectral component. This is restrictive and is only adopted for computational efficiency. Additionally, the coefficients for all $N$ frequencies are considered. Based on our previous observation, we note that the higher order frequency has minimal contribution in most physical systems. Therefore, we propose learn a 3-dimensional weight tensor $\mathbf K$ of size $m \times d \times d$, corresponding to the first $m$ frequencies. We hypothesize that this setup will enrich the parameter space and allow superior expressivity. Additionally, motivated from the Hammerstein integral, we also add the residual connection, $w$, followed by a non-linearity. With this setup, the spectral convolutional block in $j$th Sp$^2$GNO block takes the following form,
\begin{equation}\label{eq:fast-spectral}
    v_{j+1}(x)^{\text{spectral}}  =  \sigma \left(\mathbf{Q_m} \cdot \mathbf{K} \times_1 (\mathbf{Q_m}^\top \cdot v_j(x) + w(v_j(x))\right),
\end{equation}
where $\times_1$ denotes the mode-1 tensor-matrix product; this contracts the first mode of $\mathbf K$ with the columns of the product $\mathbf{Q_m}^\top \cdot v_j(x)$, which is the GFT of the node features $v_j(x) \in \mathbb{R}^{N \times d}$ of the $j$th Sp$^2$GNO block , where $j \in \{1,2,..,L\}$ is the depth of the Sp$^2$GNO block. Using the fast scalable spectral convolution introduced before, $\mathbf{Q_m}^\top \cdot v_j(x) \in \mathbb R^ {m \times d}$ contains first $m$ spectral components, corresponding to the first $m$ eigenvalues of the graph Laplacian, as it's rows. In expanded form, Eq. \eqref{eq:fast-spectral} takes the following form:
\begin{equation}\label{eq:fast-spectral1}
v_{j+1}(x)^{\text{spectral}}_{ik}  = v_j(x)^{m}_{ik}= \sigma \left(\sum_{p=1}^{m} \sum_{l=1}^{d} \sum_{n=1}^{d} \mathbf{Q_m}_{ip} \mathbf K_{pkl} \mathbf{Q_m}_{nl} v_j(x)_{in} + w(v_j(x))_{ik} \right)
\end{equation}
Note that in case the higher frequencies are active, the residual connection and the local GNN component of the overall architecture (Sec \ref{subsec:local_conv}) will compensate for the same.

\newtheorem{lemma}{Lemma}\label{lem:upperbound}
\begin{lemma}
Consider the spectral graph convolution \( \bm y = \mathbf Q g_{\bm \theta} \mathbf Q^T \mathbb X \), where \( \mathbf Q \) is the \( N \times N \) matrix of eigenvectors of the graph Laplacian, and \( g_\theta \) is a diagonal matrix representing the spectral filter. Let \( \bm{y_m} = \mathbf{Q_m} g_{\bm \theta}^{(m)} \mathbf{Q_m^T} \mathbb X \) be the approximation of the convolution using the first \( m \) eigenvectors, where \( \mathbf{Q_m} \) is an \( N \times m \) matrix containing the first \( m \) eigenvectors. Then, the relative error is bounded by:
\[
\frac{\| \bm y - \bm{y_m} \|}{\| \bm y \|} \leq \frac{\max_{i=m+1, \ldots, N} |g_{\bm \theta}(\lambda_i)|}{\max_{i=1, \ldots, N} |g_{\bm \theta}(\lambda_i)|}.
\]
\end{lemma}

\begin{proof}
Given the exact spectral graph convolution \( \bm y = \mathbf Q g_\theta \mathbf{Q^T} \mathbb X \) and its approximation \( \mathbf{Q_m} g_{\bm \theta}^{(m)} \mathbf{Q_m^T} \mathbb X\), we can express the error as:
\[
\bm e = \bm y - \bm{y_m} = \mathbf Q g_{\bm \theta} \mathbf{Q^T} \mathbb X - \mathbf{Q_m} g_{\bm \theta}^{(m)} \mathbf{Q_m^T} \mathbb X.
\]

Decompose \( \mathbf{Q} \) and \( g_\theta \) as:
\[
\mathbf{Q} = \begin{bmatrix} \mathbf{Q_m} & \mathbf{Q_r} \end{bmatrix}, \quad g_{\bm \theta} = \begin{bmatrix} g_{\bm \theta}^{(m)} & 0 \\ 0 & g_{\bm \theta}^{(r)} \end{bmatrix},
\]
where \(\mathbf{Q_r}\) contains the remaining \( N-m \) eigenvectors and \( g_{\bm \theta}^{(r)} \) contains the corresponding eigenvalues.

Substituting these decompositions, we have:
\[
\bm e = \begin{bmatrix} \mathbf{Q_m} & \mathbf{Q_r} \end{bmatrix} \begin{bmatrix} g_{\bm \theta}^{(m)} & 0 \\ 0 & g_{\bm \theta}^{(r)} \end{bmatrix} \begin{bmatrix} \mathbf{Q_m^T} \\ \mathbf{Q_r^T} \end{bmatrix} \mathbb{X} - \mathbf{Q_m} g_{\bm \theta}^{(m)} \mathbf{Q_m^T} \mathbb X.
\]

Simplifying, we get:
\[
e = \mathbf{Q_m} g_{\bm \theta}^{(m)} \mathbf{Q_m^T} \mathbb X + \mathbf{Q_r} g_{\bm \theta}^{(r)} \mathbf{Q_m^T} \mathbb X - \mathbf{Q_m} g_{\bm \theta}^{(m)} \mathbf{Q_m^T} \mathbb X = \mathbf{Q_r} g_{\bm \theta}^{(r)} \mathbf{Q_r^T} \mathbb X.
\]

Taking the norm of the error:
\[
\| \bm e \| = \| \mathbf{Q_r} g_{\bm \theta}^{(r)} \mathbf{Q_r^T} \|.
\]

Since \( \mathbf{Q_r} \) is orthogonal, \(\| \mathbf{Q_r} \| = 1\) and \( \| \mathbf{Q_r^T} \mathbb X \| \leq \| \mathbb X \| \). Therefore:
\[
\| \bm e \| \leq \| g_{\bm \theta}^{(r)} \| \| \mathbb X \|.
\]

The norm of the exact convolution is:
\[
\| \bm y \| = \| \mathbf{Q} g_{\bm \theta} \mathbf{Q^T} \mathbb X \| = \| g_{\bm \theta} \mathbf{Q^T} \mathbb X \| \approx g_{\theta,\max} \| \mathbb X \|,
\]
where \( g_{\theta,\max} = \max_{i=1, \ldots, N} |g_\theta(\lambda_i)| \).

Thus, the relative error is:
\[
\frac{\| \bm e \|}{\| \bm y \|} \leq \frac{\| g_{\bm \theta}^{(r)} \| \| \mathbb X \|}{g_{\bm \theta,\max} \| \mathbb{X} \|} = \frac{\| g_\theta^{(r)} \|}{g_{\theta,\max}}.
\]

Since \( \| g_{\bm \theta}^{(r)} \| = \max_{i=m+1, \ldots, N} |g_\theta(\lambda_i)| \), the relative error bound is:
\[
\frac{\|\bm y - \bm{y_m}  \|}{\| \bm y \|} \leq \frac{\max_{i=m+1, \ldots, N} |g_{\bm \theta}(\lambda_i)|}{\max_{i=1, \ldots, N} |g_{\bm \theta}(\lambda_i)|}.
\]
\end{proof}

\subsubsection{Local Convolution through Spatial GNN with Gating Mechanism}\label{subsec:local_conv}
A second component within the proposed approach is the local convolution achieved through spatial GNN. In equation \eqref{eq:spatial_conv}, we have defined a generalized version of spatial convolution as $\bm y = f(\mathbf A) \mathbb X$. Within the Sp$^2$GNO block, the spectral GNN will already capture the long-range dependencies; here, we only consider aggregating the local neighborhood information, enabling the spatial GNN to capture the short-range dependencies effectively. Hence, for each node, we aggregate only the node features from its 1-hop direct neighbors, such that
\begin{equation}
    f(\mathbf A) = \mathbf{A}^1 = \mathbf{A},
\end{equation}
With this, the aggregation scheme in the spatial GNN inside $j$th Sp$^2$GNO block is expressed as $v_j(x)^{\text{spatial}} = \mathbf A v_j(x) \mathbf W$, where, $v_j(x)^{\text{spatial}}$ is the output node features of the spatial GNN within $j$th Sp$^2$GNO block. Specifically for a particular node $u$ this operation corresponds to,
\begin{equation}
v_{j+1}(x)^{\text{spatial}}_u = \sum_{v \in \mathcal{N}(u)} \mathbf{W} v_j(x)_v,
\end{equation}
where $v_{j+1}(x)_u^{\text{spatial}}$ is the output node features of the spatial GNN for node $u$ within $(j+1)-$th Sp$^2$GNO block, $\mathcal{N}(u)$ denotes the neighborhood of the node $u$ and $\mathbf{W} $ is the weight for the linear transformation of the node features. We note that for the expected performance of this component, it is necessary to determine the optimal graph structure. Further details on this are provided next.
% Apart from the global convolution via spectral GNN, we also parallelly approximate the kernel integration of the Eq. \eqref{eq:vt_equation} via local convolution through a spatial GNN. In equation \eqref{eq:spatial_conv}, we have defined a generalized version of spatial convolution as $\bm y = f(\mathbf A) \mathbb X$. Within the Sp$^2$GNO block, the spectral GNN will already capture the long-range dependencies; here, we only consider aggregating the local neighborhood information, enabling the spatial GNN to capture the short-range dependencies effectively, which its spectral counterpart might miss. Hence, for each node, we aggregate only the node features from its 1-hop direct neighbors. Thus, in our case, the function $f(\mathbf A) = \mathbf{A}^1 = \mathbf{A} $. Thus, the aggregation scheme in the spatial GNN is expressed as $y = \mathbf A \mathbb X \mathbf W$. Specifically for a particular node $u$ this operation is,
% \begin{equation}
% \bm{y_u} = \sum_{v \in \mathcal{N}(u)} \mathbf{W} \mathbb{X}_v
% \end{equation}
% where $\mathcal{N}(u)$ denotes the neighborhood of the node $u$, and $\mathbf{W} $ is the weight for the linear transformation of the node features.

\noindent \textbf{Learning of the Optimal Graph Structure through Gating Mechanism:} As the initial graph structure was constructed using the k-nearest neighbors method, the connectivity of the graph is suboptimal. Thus, inspired by \cite{Gupta2023FrigateFS}, we use a gating mechanism over positional Lipshitz embeddings of the graph to effectively learn the edge weights of the k-nearest neighbor graph. Initially, we construct the graph with a relatively large value of $k$ so that the network can learn the important edges, assigning them weights close to one, and nullify the effect of spurious edges by assigning them small weights. This gating mechanism enables Sp$^2$GNO to learn the optimal graph structure in an efficient way.

% \textbf{Gating mechanism :}
Within the Gating mechanism, to calculate the edge-weight $\gamma_{uv}$ for the edge between the node $u$ and node $v$, we first assign the initial weight of the edge between any pair of nodes $u$ and $v$ to be the Euclidean distance,
\begin{equation}
    w_{uv} = \|\bm{x}_u - \bm{x}_v\|,
\end{equation}
where, $\bm{x_u}$ and $\bm{x_v}$ are the coordinates of the positions of node $u$ and node $v$, respectively. We encode these edge weights through a linear layer with weights $\mathbf{W_2}$, and concatenate the encoded edge weights with the positional embeddings of node $u$ and $v$. Finally, we pass the concatenated vector through a neural network with a sigmoid activation function in the output layer to obtain the final edge-weight $\gamma_{uv}$. Thus, the modified form of the spatial GNN inside $j$th Sp$^2$GNO block is represented as,
\begin{equation}
    v_{j+1}(x)_u^{\text{spatial}} = \sum_{v \in \mathcal{N}(u)} \gamma_{uv} \mathbf W v_j(x)_v
\end{equation}
\begin{align}
\gamma_{uv} &=
\begin{cases} 
      \mathbf \sigma_2(\mathbf{W_3}\sigma_1(\mathbf{W_1}[\bm{h_v}||\bm{h_u} || \mathbf{W_2}w_{uv}])) \ & \text{ if node } u \text{ and node } v \text{ are connected} \\
      0 & \text{otherwise} 
\end{cases}
\end{align}
where $v_j(x)_u^{\text{spatial}}$ is the feature vector for node $u$ after spatial convolution; $\bm{h_u}$ and $\bm{h_v}$ are the Lipshitz embeddings of node $u$ and node $v$, and $\mathbf{W_1}$, $\mathbf{W_3}$ are the weights of the neural network, and $\mathbf{W_2}$ are weights for linear layer. Here, $\sigma_1$ is the ReLU activation function and $\sigma_2$ is the sigmoid activation function. In compact matrix notation, this layer can be represented as,
\begin{align}
v_{j+1}(x)^{\text{spatial}} &= \Gamma \odot \mathbf A v_j(x) \mathbf W \quad \text{where} \quad \Gamma = [\gamma_{uv}] \\
\gamma_{uv} &=
\begin{cases} 
      \mathbf \sigma_2(\mathbf{W_3}\sigma_1(\mathbf{W_1}[\bm{h_v}||\bm{h_u} || \mathbf{W_2}w_{uv}])) \ & \text{ if node } u \text{ and node } v \text{ are connected} \\
      0 & \text{otherwise} 
\end{cases}
\end{align}

\noindent \textbf{Lipschitz Positional Embedding}: 
Positional Embedding (PE) is a way to construct a $n$ dimensional embedding for each node in a graph, which contains the information on the position of the node within the graph topology. We adopt Lipschitz embedding to calculate PE, which we use as inputs of the gating mechanism in the spatial GNN. In Lipschitz embedding, a set of $n$ number of nodes called "anchors" are selected from the vertex set $\mathbb V$ of the graph. Lipschitz embedding vector for a node $u$ is the $n$ dimensional vector containing the shortest path distances of the node $u$ from each node of the anchor set.

\noindent \textbf{Definition 1} (Lipschitz embedding) \textit{ Given the anchor set $\mathcal{V}_{a}=\{v_1,v_2,...,v_m\}$, the $m$ dimensional vector consisting of the shortest path distances of a particular node $u$ from each of the anchor nodes is called the Lipschitz embedding vector for the node $u$. Which is essentially}
\begin{equation}
    \bm{h_u} = \left(d( u, v_1), d(u,v_2), ..., d(u,v_n)\right) \in \mathbb{R}^n
\end{equation}
where, $d( u, v_i)$ are the shortest path distances from node $u$ to anchor node $v_i$. We decide the number of anchors $n$ in a graph with the help of the Bourgain Embedding theorem \cite{Linialgraph}. Bourgain Theorem provides the upper bound of the number of anchors $n = O(log(N)^2) $ to achieve a maximum Euclidean distortion $\alpha =O(log(N))$, where $N$ is the number of nodes in the graph. Here, $\alpha$ is the measure of how close the embedded space $Y$ is to be a subset of Euclidean space. Lower the $\alpha$, closer the $Y$ from being a subset of Euclidean space.

% \textbf{Theorem 1}(Bourgain Embedding Theorem) \textit{If there exists a mapping $f : X \to Y$ from n-point metric space $(X, d_X)$ to $(Y, d_Y)$, then to embed $X$ to $Y$ via mapping $f$, the dimension m should be at most $O(log(n)^2)$ to achieve an Euclidean distortion $O(log(n))$.}

% \newtheorem{theorem}{Theorem}\label{th:bourgain_theorem}
% \begin{theorem}[Bourgain Embedding Theorem]
%     If there exists a mapping \( \mathcal{F} : X \to Y \) from an \( N \)-point metric space \((X, d_X)\) to \((Y, d_Y)\), then to embed \( X \) into \( Y \) via the mapping \( \mathcal{F} \), the dimension \( n \) should be at most \( O(\log(N)^2) \) to achieve a Euclidean distortion of \( O(\log(N)) \).
% \end{theorem}

\begin{theorem}[Bourgain Embedding Theorem]
\label{th:bourgain_theorem}
    If there exists a mapping \( \mathcal{F} : X \to Y \) from an \( N \)-point metric space \((X, d_X)\) to \((Y, d_Y)\), then to embed \( X \) into \( Y \) via the mapping \( \mathcal{F} \), the dimension \( n \) should be at most \( O(\log(N)^2) \) to achieve a Euclidean distortion of \( O(\log(N)) \).
\end{theorem}

The proof of the Theorem \ref{th:bourgain_theorem} can be found in \cite{Linialgraph}. In our case, the mapping $\mathcal{F}$ is nothing but the Lipschitz embedding defined on metric space $X$, which is our graph $G$ with $N$ nodes. The distance metric $d_X $ is $d(u,v)$, where $d(u,v)$ is the shortest path distance between the nodes $u$ and $v$. To calculate the Lipschitz embedding of a graph with $N$ nodes, we select the number of anchor nodes to be $n \approx log(N)^2$.
\begin{algorithm}[ht!]
\caption{Algorithm of the Sp$^2$GNO}
\label{alg:sp2gno}

\begin{algorithmic}[1]
\REQUIRE $N$-samples of the pair $\{a(x) \in \mathbb{R}^{n_D \times d_a}, u(x) \in \mathbb{R}^{n_D \times d_u} \}$, coordinates $x \in D$, and network hyperparameters.
\STATE Concatenate the inputs: $\left[a(x) || x \right] \in \mathbb{R}^{n_D \times (d_a +d)}$.
\STATE Calculate the adjacency matrix using KNN method : $A \in \mathbb{R}^{N \times N}$
\STATE Calculate the degree matrix : $\mathbf{D} : D_{ii} = \sum_{j} A_{ij}$
\STATE Calculate the normalized graph Laplacian : $\mathbf{\hat{L}} = I - \mathbf{D}^{-1/2} \mathbf{A}\mathbf{D}^{-1/2}$
\STATE Calculate the first $m$ eigenvectors $\mathbf{Q_m} \in \mathbb{R}^{N \times m}$ of $\mathbf{\hat{L}}$ using LOBPCG alogorithm.
\STATE Calculate the Lipschitz positional embeddings $\bm{h} \in \mathbb{R}^{N \times n}$ of all the nodes using $n$ anchors. 
\FOR{epoch $=$ 1 to epochs}
    \STATE Uplift the input using transformation $P(\cdot)$: $v_0(x) \in \mathbb{R}^{N \times d_v} = P(\left[a(x)|| x\right])$.
    \FOR{$j = 0$ to $L-1$ perform the iterations: $ v_{j+1} = {S_{j+1}(v_j)}$}

        \STATE Perform the spectral convolution: $v_{j+1}(x)^{\text{spectral}} = \sigma \left(\mathbf{Q_m} \cdot \mathbf{K} \times_1 \mathbf{Q_m}^\top \cdot v_j(x) + w(v_j(x))\right)$
        \STATE Perform the Spatial Convolution: $v_{j+1}(x)^{\text{spatial}} = \Gamma \odot \mathbf A v_j(x) \mathbf W$\\
        where $\Gamma = [\gamma_{uv}]$ and $ \mathbf \gamma_{uv} =\sigma_2(\mathbf{W_3}\sigma_1(\mathbf{W_1}[\bm{h_v}||\bm{h_u} || \mathbf{W_2}w_{uv}]))$ if there is an edge between $u$, and $v$, else $\gamma_{uv} = 0$
        \STATE Concatenate outputs of both convolution: $ y = \left[v_{j+1}(x)^{\text{spectral}}|| v_{j+1}(x)^{\text{spatial}}\right]$ 
        \STATE Calculate the output using $f(\cdot)$: $v_{(j+1)}(x) = f(y)$
    
    \ENDFOR
    
    \STATE Compute the final output using : $\hat{u}(x) \in \mathbb{R}^{N \times d_{u}} = Q(v_L(x))$ \hfill \textit{($Q(\cdot) : \mathbb{R}^{d} \rightarrow \mathbb{R}^{d_u}$ is a FNN)}
    \STATE Compute the loss: $L(u, \hat{u}; \bm \theta)$.
    \STATE Compute the gradient of the loss: $\frac{\partial L}{\partial \bm \theta_{\text{NN}}} (u, \hat{u}; \bm \theta_{\text{NN}})$.
    \STATE Update the parameters of the network using the gradient.
\ENDFOR
% \ENSURE Solution space $u \in U$, parameters of NN $\theta_{\text{NN}}$.
\RETURN Parameters of NN $\bm \theta_{\text{NN}}$
\end{algorithmic}
\end{algorithm}
\subsection{Collaboration between Spatial and Spectral GNN}
After being processed by both spatial and spectral GNNs, we concatenate outputs of both and pass it through a function $f : \mathbb{R}^{2d} \to  \mathbb{R}^{d}$ to give the final output of the Sp$^2$GNO layer. This function is approximated by a linear layer, which decides the weights of contributions of both the GNNs in the final output for optimal learning. This ensures that the final output has both long-range and short-range information from both GNNs in appropriate proportion. This facilitates optimal learning without encountering the over-smoothing issue. We only learn the spectral filter corresponds to the first $m$ frequency in spectral GNN. Collaboration also ensures that if any frequency other than the lowest $m$ frequencies is present, the spatial GNN will capture it. In a nutshell, the operation within the $j$th Sp$^2$GNO block is given by,
\begin{align}
v_{j+1}^{\text{spatial}} &= \mathbf \Gamma \odot \mathbf A v_j(x) \mathbf W \quad  \\
v_{j+1}^{\text{spectral}} &= \sigma \left(\mathbf{Q_m} \cdot \mathbf K \times_1 \mathbf{Q_m}^\top \cdot v_j(x) + w(v_j(x)) \right) \\
v_{j+1}(x) &= f([\bm{v_{j+1}^{\text{spatial}}} \| \bm{v_{j+1}^{\text{spectral}}} ])
\end{align}
where $\mathbf \Gamma$ is an $N \times N$ matrix with its elements 
\begin{equation}
    \gamma_{uv} = \left\{ \begin{array}{ll}
         \sigma_2( \mathbf{W_3}\sigma_1 ( \mathbf{W_1}[h_v \| h_u \| \mathbf{W_2}w_{uv}])) & \text{if $u$ and $v$ are connected}  \\
         0 & \text{elsewhere}
    \end{array}
    \right.
\end{equation}
In above equations, $\odot$ represents the Hadamard product, and $||$ represents the concatenation operation.

The overall algorithm for training the proposed Sp$^2$GNO is shown in Algorithm \ref{alg:sp2gno}.

\section{Numerical experiments}
\label{sec:Experiments}
In this section, we present four canonical examples representing various classes of problems in solid and fluid mechanics to illustrate the performance of the proposed approach. 
For all the examples, we treat the available data as point clouds and construct the initial graph structure by using the $k-$nearest neighbor (KNN) algorithm. For all the problems, we have considered {$l=6$ Sp$^2$GNO blocks and \texttt{GELU} activation operator. A shallow neural network is used as the uplifting layer, with the uplifted dimension being 32. A 2-layer deep neural network has been used for the downlifting layer. 
For all examples, the performance of the proposed approach is evaluated using contour plots and normalized mean-squared error (N-MSE). The results obtained are compared against state-of-the-art neural operators including the DeepONet \cite{Deeponet}, Fourier Neural Operator (or its variant) \cite{Li2022FourierNO}, Spectral Neural Operator (SNO) \cite{Fanaskov2022SpectralNO}, and Graph Neural Operator (GNO) \cite{Li2020NeuralOG}. For training the model, we employed the ADAM optimizer with an initial learning rate of $0.001$ and a weight decay of $1e-4$. The batch size and other problem specific details are provided with each example.}

% optimization of the parameters of the WNO, the ADAM optimizer with an initial learning rate of ?? and a weight decay of 10−6 is utilized. During the optimization, decay in the learning rate of each parameter group is introduced at every 50 epochs at a rate of 0.75. The total number of epochs used for training the WNO architecture against the undertaken examples are given in 
% Table 2
% . The batch size in the data loader varies according to the underlying system between 10–25. 

% We test the performance Sp$^2$GNO in various PDE solving tasks by making it learn the solution operator of the PDE from data. We show that Sp$^2$GNO performs as well as SOTA or outperforms SOTA without any restriction on the domain geometry. It also handles irregularity in mesh structure effectively making the model suitable for a wide range of problems.

\subsection{Darcy Flow Problem}
As the first example, we consider the 2D darcy flow problem in unit square grid, which is a second-order elliptic PDE. 
\begin{equation}
\begin{aligned}
-\nabla \cdot \left( a(x) \nabla u(x) \right) &= f(x) \quad \text{for } x \in (0, 1)^2 \\
u(x) &= 0 \quad \text{for } x \in \partial (0, 1)^2
\end{aligned}
\end{equation}
% where, $a(x)$ is the permeability and $f$ is the forcing function and the dependent variable $u(x)$ is pressure. The Darcy flow problem has numerous application in fluid mechanics including modelling the flow of a fluid in a porous medium like flow of water in sand, modelling the pressure of a subsurface flow, modelling the .
% The Darcy flow equation, mathematically articulated as
% \begin{equation}
% - \nabla \cdot \left( a(x) \nabla u(x) \right) = f(x), \quad x \in (0, 1)^2,
% \end{equation}
This equation represents fluid movement through a porous medium. Here, \( u(x) \) denotes the pressure field, \( a(x) \) signifies the permeability of the medium, and \( f(x) \) serves as a source term that accounts for any sinks or sources within the domain. The unit square \( (0, 1)^2 \) is the spatial domain under consideration, and the boundary condition
\begin{equation}
u(x) = 0, \quad x \in \partial (0, 1)^2,
\end{equation}
imposes that the pressure is zero along the boundary, a common scenario encountered in practical applications. This equation is paramount in hydro-geology for modeling groundwater flow, modeling the flow of a fluid in a porous medium like the flow of water in sand \cite{Bear1984FundamentalsOT}. 
% In elasticity, this equation is used to model deformation in linear elastic materials\cite{Antman1994NonlinearPO}. 
For example, it is used in petroleum engineering for simulating the extraction of hydrocarbons, and in environmental engineering for understanding contaminant transport in soil. The permeability \( a(x) \) often exhibits spatial variability, reflecting the heterogeneous nature of real porous media, which significantly influences the flow dynamics.
Solving the Darcy flow equation analytically is usually infeasible due to the complexity introduced by the spatial variability of \( a(x) \) and \( f(x) \). Naturally, numerical simulations are commonly employed for obtaining approximate solutions. However, as the permeability field changes, one needs to re-run a numerical simulation from scratch; this makes any process involving multiple run computationally expensive. Therefore, the objective here is to train the proposed approach to learn the mapping between the input \(a(x)\) and the response \(u(x)\), $F_\theta : [x,a(x)] \to u(x) $.

\noindent \textbf{Experimental Setup} For this example, we consider that data is available to us as a point cloud with over 6500 points. To obtain the graph structure, we consider each point in the point cloud to be a node and connect it with $k= 20$ nearest nodes by using the KNN algorithm as mentioned before. The data used corresponds to Darcy flow data on a unit square domain $x \in (0,1)^2$ with $f(x) = 1$. For training the Sp$^2$GNO model, we have considered $m=32$; this indicates that only the first 32 dominant frequencies are retained. Other setups remain the same as discussed before.

% For the Darcy flow problem, we use the 2 D Darcy Flow data with resolution 421 provided by FNO \cite{li2020fourier}. The data was generated in a unit square domain $x \in (0,1)^2$ with Dirichlet boundary condition $u(x) = 0$. The source term was considered to be unity during data generation, i.e., $f(x) = 1$. For given coefficient values $a(x)$ and coordinate values $x$  as input, we train our model to predict the solution $u(x)$. Thus, we learn the operator $F_\theta : [x,a(x)] \to u(x) $. We report \textit{Normalised Mean Squared Error} (N-MSE) as a metric for measuring model performance. We use the first 32 lowest Graph frequencies for learning. We train all the models on a resolution of $85$.
% As baselines, we use FNO \cite{li2020fourier}, UFNO \cite{Wen2021UFNOA}, DeepONet \cite{Deeponet}, SNO \cite{Fanaskov2022SpectralNO}, and GNO \cite{Li2020NeuralOG}.

\begin{figure}[ht]
  \centering
  \includegraphics[width=1.0\textwidth]{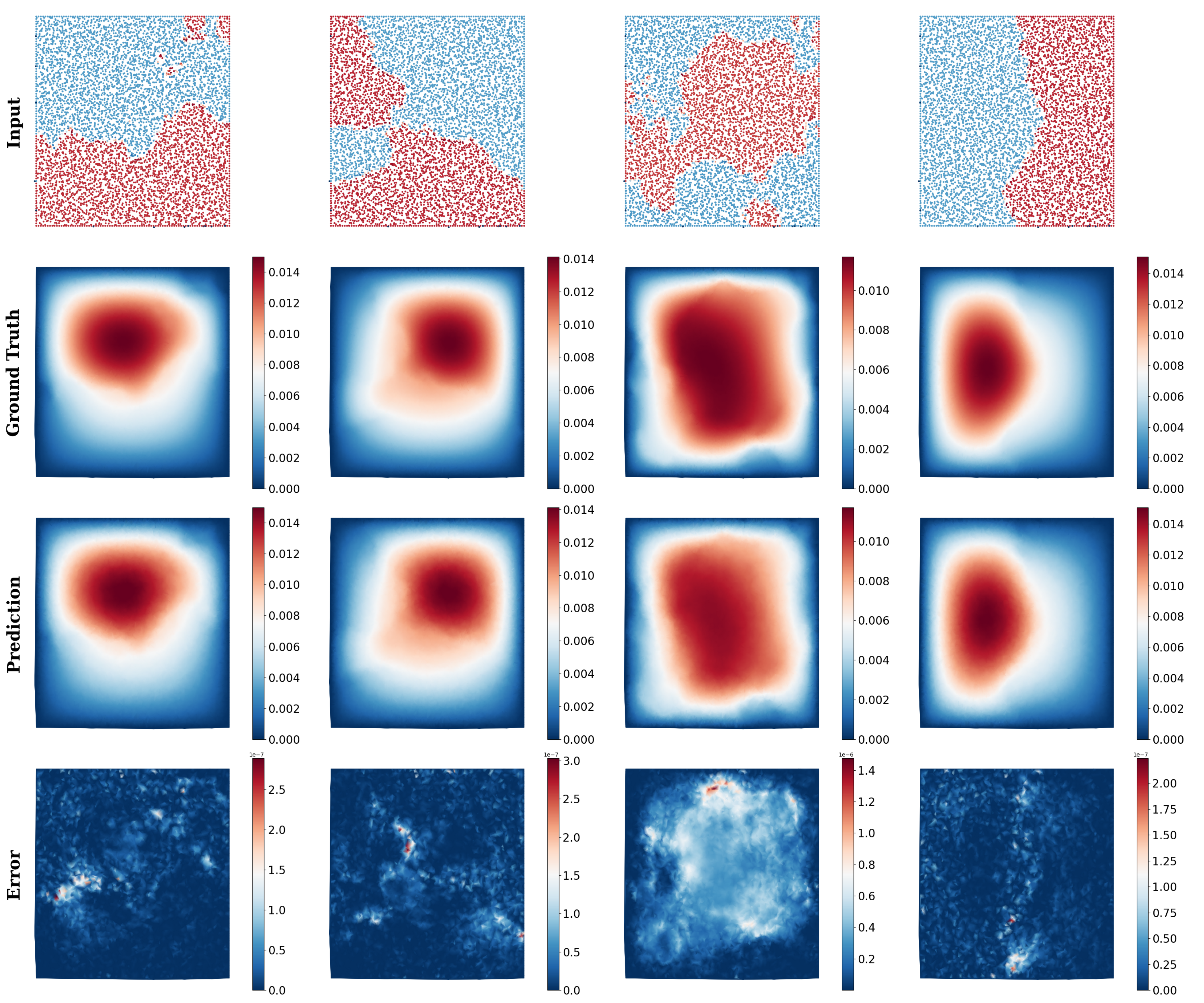}
  \caption{\textbf{Predictions in Darcy dataset}: We have plotted input coefficients, ground truth, predictions, and errors of four different test examples in four columns. The first row represents the permeability coefficients for each of the four examples. The second and third rows represent the ground truth and predictions for the same. The fourth row shows the MSE error.}
  \label{fig:darcy}
\end{figure}
\begin{figure}[ht]
  \centering
  \includegraphics[width=0.85\textwidth]{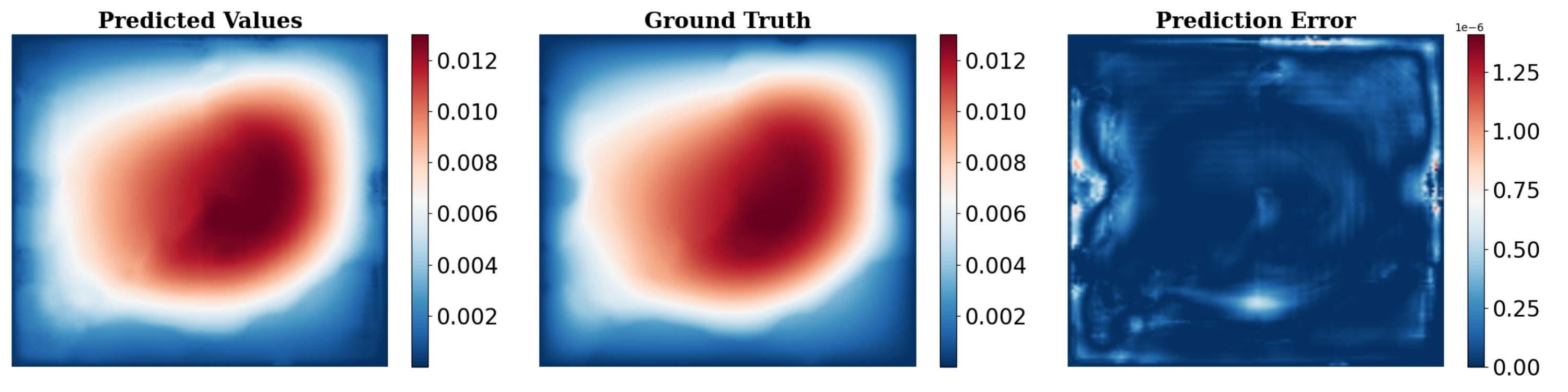}
  \caption{\textbf{Super resolution Predictions in Darcy dataset}: The leftmost figure is the prediction in a $141 \times 141$ resolution. The figure in the middle is the ground truth in $85 \times 85$ resolution. The figure on the right-hand side is the MSE error.}
  \label{fig:super_resolution}
\end{figure}

\noindent \textbf{Results and discussion} The pressure contours obtained using the proposed approach is shown in Fig. \ref{fig:darcy}. Results corresponding to four different $a(x)$ are shown. For all the cases, we observe that the results predicted using the proposed Sp$^2$GNO match almost exactly with the ground truth data obtained from numerical simulation. We further carry out a comparative study with other operator learning algorithms existing in the literature. However, many of the existing operator learning algorithms cannot accommodate unstructured data. Accordingly, {for this part of the study, we consider structured data from \cite{li2020fourier} on a $85 \times 85$ grid.} We observe that the proposed approach yields the best result with relative MSE of $9.0 \times 10^{-3}$, followed by FNO variants and GNO. We also illustrate that the trained model can be used to predict at other resolutions without any retraining. Here we used the trained model to predict the response for a point cloud with approximately 20,000 points. The resulting contours are shown in Fig. \ref{fig:super_resolution}. In this case also, the results obtained using the trained model shows a good match with the ground truth obtained using numerical simulation. 

% the resolution invariance of the proposed approach. 

% \textbf{Discretization Invariance and Super-Resolution}   Discretization Invariance \cite{li2020fourier} is the core property of an operator, which makes it an efficient algorithm against the traditional numerical solvers. This means that an operator can be trained in any arbitrary resolution and tested for any different discretization and resolution. The same set of parameters during training can transferred while testing, thus there is no requirement of retraining the model to get predictions in different resolutions. Numerical simulations are very expensive when it comes to solving a PDE in very fine resolutions. As a remedy, an operator architecture can be trained in inexpensive coarse-grained low-resolution data from numerical solvers and can predict in very high resolution without any significant loss of accuracy, enabling an operator capable of Super Resolution. To test the super Resolution capability of Sp$^2$GNO, we trained the model in the Darcy dataset in $85 \times 85$ resolution and tested it in $141 \times 141$ resolution. The training error in $85 \times 85$ resolution was $0.0113$, and the test error in $141 \times 141$ resolution was $0.0333$, which proves its efficient Super Resolution capability. In figure \ref{fig:super_resolution}, we show the contour plots of the predicted values, ground truth, and MSE prediction errors.

% \textbf{Results} we train all the models for $3000$ epochs. From the table \ref{table:performance_comparison}, we can see that Sp$^2$GNO achieves the lowest relative MSE of $0.0095$ on the test data among all the SOTA models. The performance of DeepONet and SNO is not so good in the Darcy dataset, while they give a test use of $0.0588$ and $0.049$, respectively.  The second best model in this dataset is FNO, which achieves a relative test error of $0.0108$. Thus, Sp$^2$GNO achieves an improvement of 12 percent over FNO. The contour plots of predictions are shown in fig \ref{fig:darcy}. We can see the predictions and ground truths are almost identical in the contour plots.

% \begin{figure}[h]
%   \centering
%   \includegraphics[width=1.0\textwidth]{Final_plots/darcy_at_epoch_2980_loss_0.0095.png}
%   \caption{\textbf{Predictions in Darcy dataset}: We have plotted ground truth, predictions and errors of 5 different test examples in five columns. The first row represents the ground truths for each of the five examples. The second and third rows represent the predictions and ground truth for the same.}
%   \label{fig:darcy}
% \end{figure}

\subsection{Flow Around Airfoils}
As second example, we consider a problem involving flow around airfoil. 
% Transonic flow around an airfoil is governed by the Euler equation given by,
% \begin{equation}
% \frac{\partial \rho}{\partial t} + \nabla \cdot (\rho \mathbf{v}) = 0 ,\quad 
% \frac{\partial (\rho \mathbf{v})}{\partial t} + \nabla \cdot (\rho \mathbf{v} \otimes \mathbf{v} + p\mathbf{I}) = 0 ,\quad 
% \frac{\partial E}{\partial t} + \nabla \cdot \left( (E + p) \mathbf{v} \right) = 0
% \end{equation}
% where, $\rho$ is  is the density of air, $v$ is the velocity vector, $p$ is pressure and $E$ is total energy.
The study of transonic flow over an airfoil is crucial in the field of computational fluid dynamics (CFD), and it has significant implications for aerospace engineering. The behavior of such flows is governed by the Euler equations, which leverage the conservation principles of mass, momentum, and energy for a non-viscous, compressible fluid. These equations are mathematically expressed as follows:
\begin{align}
    \frac{\partial \rho}{\partial t} + \nabla \cdot (\rho \bm{v}) &= 0, \label{eq:continuity} \\
    \frac{\partial (\rho \bm{v})}{\partial t} + \nabla \cdot (\rho \bm{v} \otimes \bm{v} + p \mathbf{I}) &= 0, \label{eq:momentum} \\
    \frac{\partial E}{\partial t} + \nabla \cdot ((E + p) \mathbf{v}) &= 0, \label{eq:energy}
\end{align}
where $\rho$ denotes the fluid density, $\bm{v}$ is the velocity vector, $p$ represents the pressure, and $E$ is the total energy. For our problem setup, the far-field boundary conditions are specified with a fluid density $\rho_\infty = 1$, pressure $p_\infty = 1.0$, Mach number $M_\infty = 0.8$, and an angle of attack (AoA) set to $0^\circ$. At the airfoil surface, a no-penetration condition is enforced, ensuring that the velocity component normal to the surface is zero, thus accurately modeling the impermeability of the solid boundary. Flow modeling around airfoils plays an important role in aerospace engineering.  This requires solving the Euler equations in Eqs.\eqref{eq:continuity} -- \eqref{eq:energy}. Analytical solutions to these equations are not available for most real-world scenarios and hence, numerical solutions are obtained by employing numerical techniques such as FEM and FVM. As the shape of the airfoil changes, one needs to recalculate the solution of these equations from scratch, which makes the overall process computationally expensive. Thus, we train our proposed architecture for fast inference of the Mach number $u(x)$ from the mesh coordinates $x$. 
Here, the objective is to learn the operator from mesh coordinates $x$ to the response  Mach-number $u(x)$, $F_\theta : x \to y(x)$.

\noindent \textbf{Experimental Setup}  We use the airfoil data introduced in \cite{Li2022FourierNO}, which has 2490 pairs of input and output data each having about 11000 points. 
%
% in 
% $221 \times 51$ structured mesh points.
% The data was generated using the far-field boundary conditions $\rho_\infty = 1$, $p_\infty = 1$, $M_\infty = 0.8$ and angle of attack was fixed at zero. Here, $M_\infty$  is the mach number. 
The dataset was generated using the NACA-0012 airfoil profile, parameterized using the design element approach as delineated by Farin \cite{farin1993curves}. This involves using control nodes to parameterize the shape of the airfoil. By displacing these control nodes in a systematic manner, various airfoil geometries are generated, which are then used as input geometries for the simulation. For each modified airfoil shape, the Euler equations are solved numerically. The far-field boundary conditions and the no-penetration condition at the airfoil surface are strictly enforced to ensure realistic flow characteristics. High-fidelity CFD solvers are employed to solve these equations, thereby producing detailed and accurate flow fields around the airfoil for each configuration.
We use $1000$ pairs for training and $200$ for testing for Sp$^2$GNO and all the baselines. We have used $k=30$ in the KNN method to obtain the graphs for each domain geometry. The batch size 20 was used during training. We have only used the first $m = 48$ graph-frequencies to train the model. Other setups remain the same as before. We compare the performance of Sp$2$GNO with SOTA architectures including FNO and its variants, DeepONet \cite{Deeponet}, SNO \cite{Fanaskov2022SpectralNO}, and GNO \cite{Li2020NeuralOG}.

\begin{figure}[ht]
  \centering
  \includegraphics[width=1.0\textwidth]{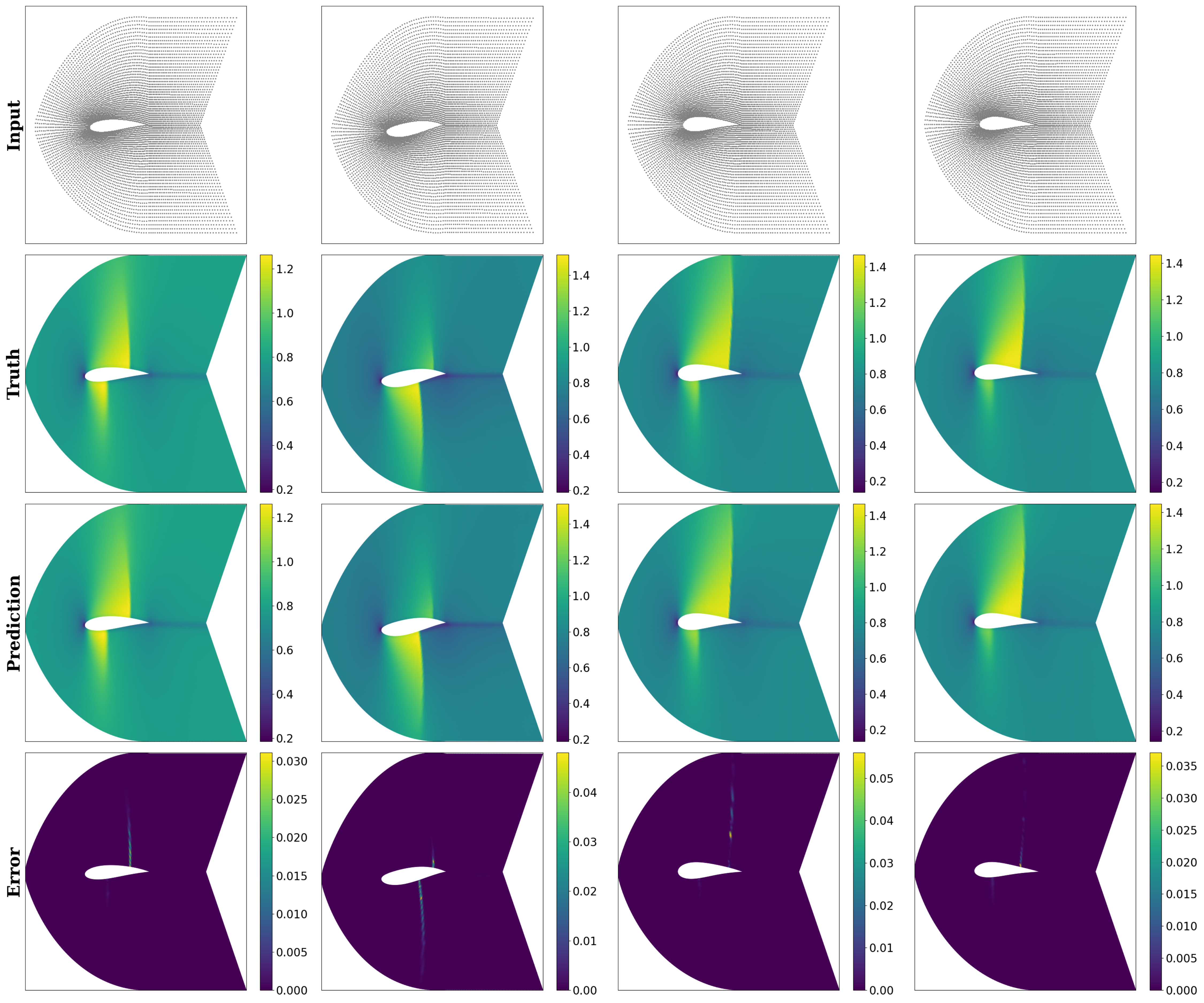}
  \caption{\textbf{Predictions in Airfoil dataset}: We have plotted input mesh coordinates, ground truth, predictions, and errors of four different test examples in four columns. The first row represents the mesh coordinates for each of the four examples, which serves as the input for the model. The second and third rows represent the ground truth and predictions for the same. The fourth row shows the MSE error.}
  \label{fig:airfoil}
\end{figure}

\noindent \textbf{Results} The contours of the Mach numbers obtained during the prediction are shown in Fig. \ref{fig:airfoil}. The contours corresponding to four different airfoil shapes as shown in the first row. For all four cases, we can see that predictions of our proposed Sp$^2$GNO almost exactly match the solution obtained from the numerical simulations, and it can effectively learn the shock. The relative test MSE for the Airfoil problem is also reported in Table \ref{table:performance_comparison}. In airfoil, the proposed Sp$^2$GNO shows the best performance with a relative error of $9.8 \times 10^{-3}$ followed by FNO and its variant, DeepONet, GNO, and SNO. Here, it should be noted that Sp$^2$GNO shows a significant performance gain of 24 percent over its nearest competitor.

\subsection{Hyper-elastic material}

As the third example, we consider elastic deformation of solid. The governing equation for this takes the following form,
\begin{equation}\label{eq:hyperelasticity}
\rho_s \frac{\partial^2 \bm{u}}{\partial t^2} + \nabla \cdot \bm \sigma = 0,
\end{equation}
where $\rho_s$ is the mass density, $\bm{u}$ represents the displacement vector, and $\bm \sigma$ denotes the stress tensor. We have considered hyperelastic material and modeled it using incompressible Rivlin-Saunders constitutive relation,
\begin{equation}
w(\epsilon) = C_1 (I_1 - 3) + C_2 (I_2 - 3),
\end{equation}
where $I_1 = \text{tr}(C)$ and $I_2 = \frac{1}{2} \left( (\text{tr}(C))^2 - \text{tr}(C^2) \right)$ are the invariants of the right Cauchy-Green stretch tensor $C = 2\epsilon + I$. The parameters for the energy density function are $C_1 = 1.863 \times 10^5$ and $C_2 = 9.79 \times 10^3$. We have considered a squared domain with a void characterized by the coordinates $\bm x$. The objective here is to compute the stress distribution for a given geometry. This can be solved by using numerical techniques such as FEM; however, as the geometry changes, one needs to run the numerical solver from scratch. Therefore, training a deep neural operator to learn the mapping between the coordinates $\bm x$ and stress components $\bm \sigma (x)$, $F_\theta: \bm x \mapsto \bm \sigma (x)$.

\noindent \textbf{Experimental Setup}  
For this problem, we consider a unit square domain with an arbitrary void at the center of the domain. 
The data is generated in a unit square domain in an irregular point cloud structure with an arbitrarily shaped void in the center of the domain. The prior distribution for the void radius $r$ is given by
\begin{equation}
r = 0.2 + \frac{1 + \exp(\tilde{r})}{0.2},
\end{equation}
with $\tilde{r} \sim N(0, 42 (-\nabla + 32)^{-1})$, ensuring the constraint $0.2 \leq r \leq 0.4$.
The unit cell is subjected to specific boundary conditions, with the bottom edge clamped and a tensile traction $\mathbf{t} = [0, 100]$ applied along the top edge. Using a finite element solver described by Huang et al. \cite{huang2010adaptive}, we simulate the system with approximately 100 quadratic quadrilateral elements per unit cell, accurately capturing the stress and strain distributions. The data generation process involves running 1000 simulations to create the training dataset and an additional 200 simulations for testing. There are $972$ grid points in data. The models were trained to predict the inner stress of the material at each point, given the position coordinates of the points. We used the KNN method with $k = 20$ to obtain the domain graph from the coordinates of domain vertices. We have used the first $m = 48$ graph frequencies in the model, and batch size $1$ was used to train the model.
We compare our results with other neural operators. GNO DeepONET and Geo-FNO can handle the irregularities in the mesh structures; thus, we train these along with Sp$^2$GNO directly in the irregular grid. The other models cannot handle the irregularities, thus we train them in an equivalent interpolated $41 \times 41$ regular grid for comparison.

\begin{figure}[ht]
  \centering
  \includegraphics[width=1.0\textwidth]{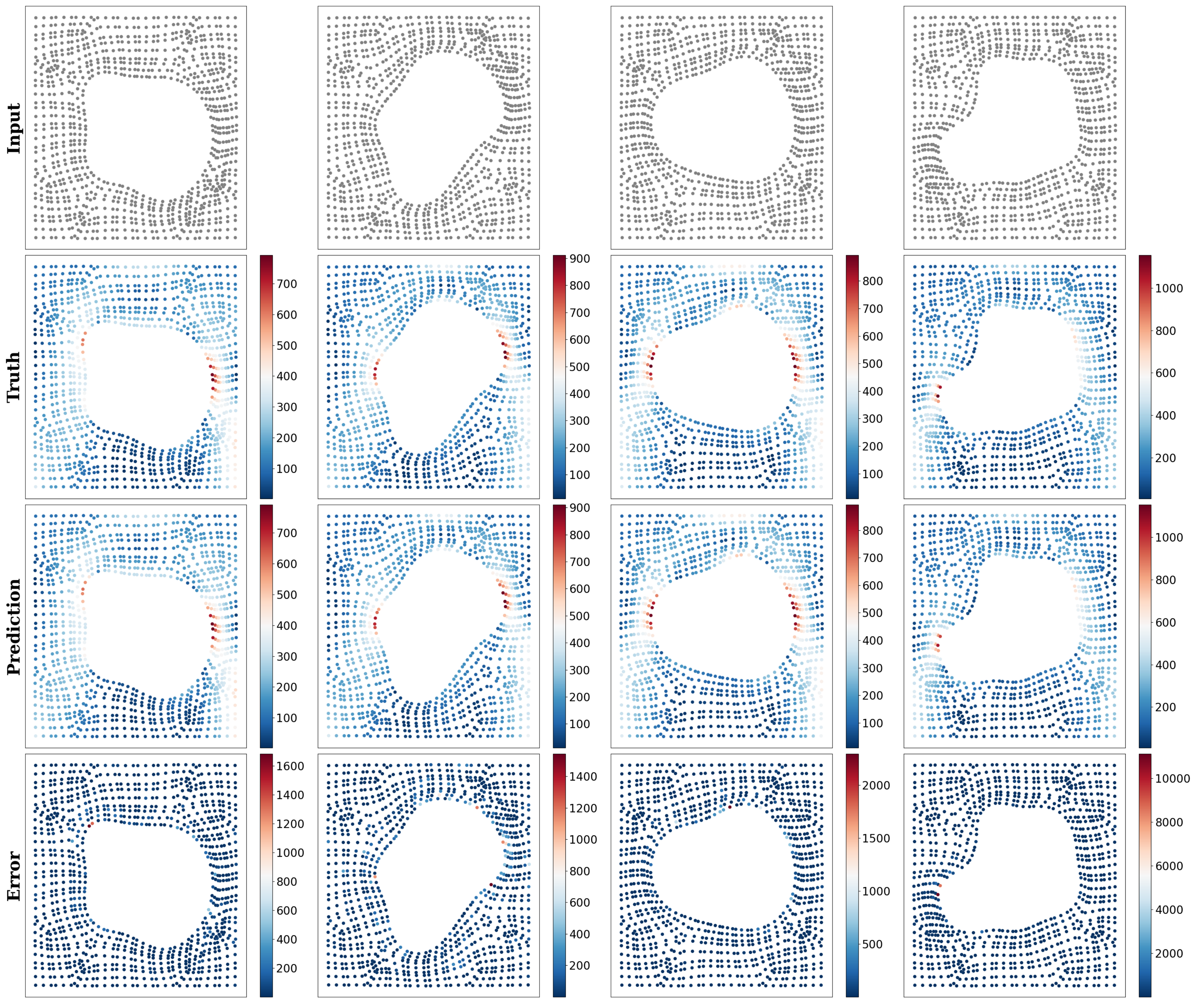}
  \caption{\textbf{Predictions in Elasticity dataset}:  We have plotted input mesh coordinates, ground truth, predictions, and errors of four different test examples in four columns. The first row represents the mesh coordinates for each of the four examples, which serves as the input for the model. The second and third rows represent the ground truth and predictions for the same. The fourth row shows the MSE error.}
  \label{fig:elasticity}
\end{figure}

\noindent \textbf{Results} The scatter plot of the obtained stress distribution is shown in Fig \ref{fig:elasticity}. We can see the predictions of the Sp$^2$GNO accurately match with the solutions obtained through numerical solvers. This proves that Sp$^2$GNO can effectively handle the irregularity in grid structure, which makes it flexible in terms of grid structure and domain geometry. Quantitatively, Geo-FNO and SP$^2$GNO yield the best results followed by SNO. The higher error corresponding to vanilla FNO can be attributed to the interpolation error.

% As FNO, U-FNO, and SNO can't handle irregularities in the domain and grid structure, we train these models in interpolated regular grid data. Thus, these models are performing relatively well on the interpolated elasticity data. FNO performs best among them with a relative MSE of $0.0229$. We train GNO and DeepONet and the proposed Sp$^2$GNO directly in the original irregular elasticity data. Among the models that were trained in irregular elasticity data, Sp$^2$GNO performs best with a relative MSE of $0.0262$. 

\subsection{Incompressible Navier-stokes Equation}
As the last example, we consider the  Navier-Stokes equation. This is fundamental in fluid mechanics, describing how fluids like air and water move, which is crucial for predicting weather, designing aircraft and vehicles, and understanding ocean currents. Mathematically, the Navier-Stokes equation is written as,
\begin{equation}\label{eq:navier_stokes}
\nabla \cdot \bm{U} = 0 \quad , \quad
\frac{\partial w}{\partial t} + \bm{U} \cdot \nabla w = \nu \nabla^2 w + f \quad ,\text{with} \quad
w|_{t=0} = w_0
\end{equation}
where, $\bm U=(u_x,u_y)$ is the velocity vector in 2 dimension, $w = |\nabla \times \bm U| = \frac{\partial u_x}{\partial y} - \frac{\partial u_y}{\partial x}$ is the vorticity. Here, we choose the $w_0$ to be the initial condition for the velocity field, and $f$ represents the forcing function of the system. The viscosity $\nu = 10^{-5}$ here is a small positive constant. The objective here is to train a neural operator to learn the dynamics of the underlying system governed by the Navier-Stokes equation. In particular, we train our proposed model to learn the solution operator for the Eq. \eqref{eq:navier_stokes}, enabling it to predict the next $10$ time-steps when data of the first $10$ time-steps is given as input.

\noindent \textbf{Experimental Setup}
For the Navier-Stokes problem, we use the 2 dimensional Navier-Stokes data with viscosity $\nu = 10^{-5}$.  The overall domain is discretized into $\approx 4000$ point clouds.
To construct the domain graph, we employ KNN method with $k = 30$. We have used the first $m =32$ graph frequencies within our model and have trained it using batch size $1$.

\begin{figure}[ht]
  \centering
  \includegraphics[width=\textwidth]{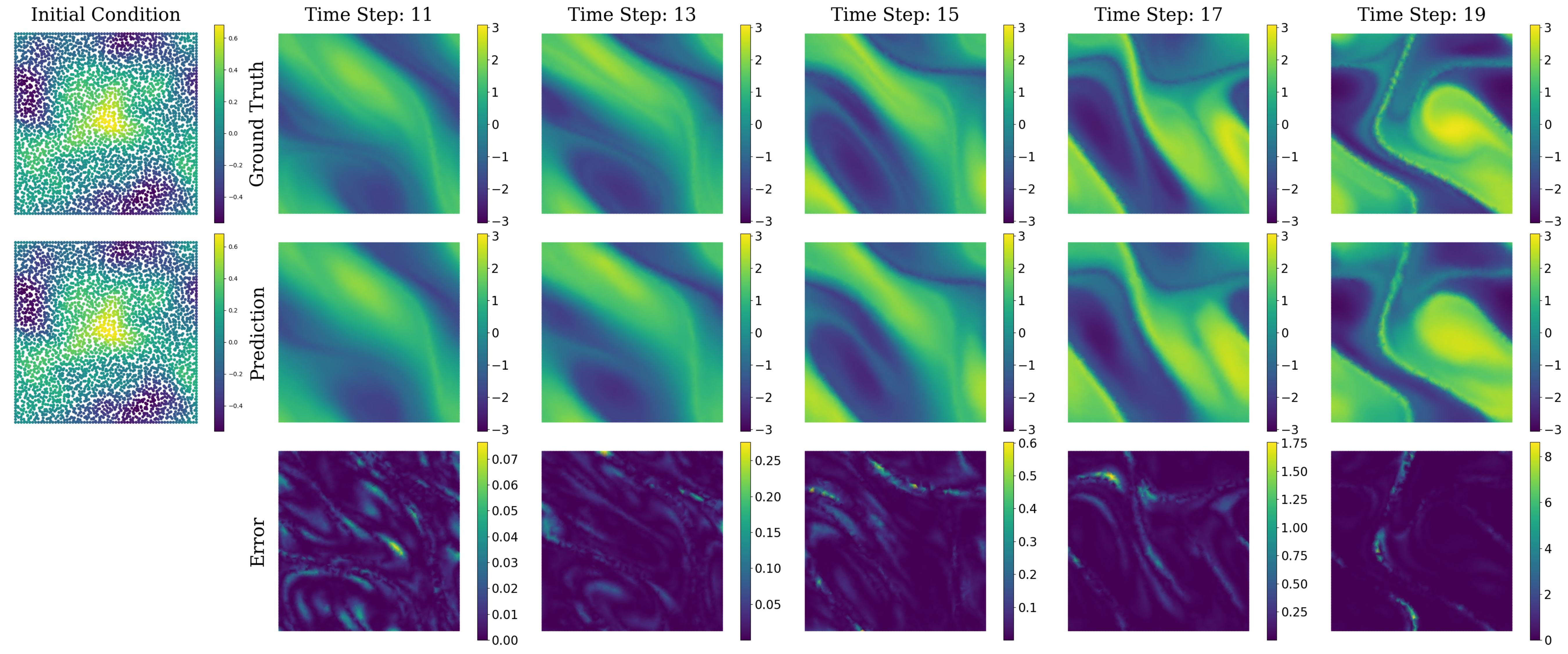}
  \caption{\textbf{Predictions in Navier-Stokes dataset}: From left to right, the first column is the initial condition. Each of the next five columns corresponds to time-step 1 to 9. The first row corresponds to the ground truth at each time step. The second and third rows display the prediction and MSE errors for each step.}
  \label{fig:navier_stokes}
\end{figure}

\noindent \textbf{Results} The contour plots obtained during the prediction on the test data is illustrated in Fig. \ref{fig:navier_stokes}. We observe that the predictions of Sp$^2$GNO are visually indistinguishable
from those obtained through numerical simulations. This illustrate the capability of the proposed approach in solving complex problems such as the ones governed by Navier-Stokes equation. To benchmark the performance of the proposed approach against other approaches, we use the unstructured data for Geo-FNO, DeepONet, and the proposed SP$^2$GNO. As for FNO and SNO, we consider the data on a regular grid (obtained through interpolation). GNO did not converge and hence, results corresponding to the same are not reported. We observe that Geo-FNO and the proposed SP$^2$GNO yield the best result followed by FNO and UFNO. 
The superior performance of Geo-FNO and SP$^2$GNO can be attributed to the fact that both these algorithms exploits original data on point cloud. The other FNO variants and SNO, on the other hand, are prone to initial interpolation error because of the interpolation. Overall, this example clearly indicates the applicability of the proposed approach in learning dynamics from point clouds in an effective manner.

\begin{table}[ht]
  \caption{Performance comparison with different baselines on selected benchmarks. MSE is recorded. A smaller MSE indicates better performance.}
  \label{table:performance_comparison}
  \centering
  \resizebox{\textwidth}{!}{
    \begin{tabular}{lccccc}
      \toprule
      \textbf{MODEL} & \textbf{HYPERELASTICITY} & \textbf{NAVIER–STOKES} & \textbf{DARCY} & \textbf{AIRFOIL}  \\
      \midrule
      % U-NET (2015) & 0.0235 & 0.1982 & 0.0080 & 0.0079 & 0.0065 \\
      % SWIN (2021) & 0.0283 & 0.2248 & 0.0397 & 0.0270 & 0.0109 \\
      DeepONet & $9.7\times10^{-2}$ & $3.0\times10^{-1}$ & $5.9\times10^{-2}$ & $3.9\times10^{-2}$  \\
      FNO & $5.1\times10^{-2}$ & $1.9\times10^{-1}$ & $1.1\times10^{-2}$ & $2.9\times10^{-2}$  \\
      Geo-FNO & $\bm{2.3\times10^{-2}}$ & $\bm{1.6\times10^{-1}}$ & $1.1\times10^{-2}$ & $1.4\times10^{-2}$  \\
      % U-FNO & 0.0239 & 0.2231 & 0.0183 & 0.0269   \\
      SNO & $3.9\times10^{-2}$ & $2.6\times10^{-1}$ & $5.0\times10^{-2}$ & $8.9\times10^{-2}$   \\
      GNO & $1.7\times10^{-1}$ & - & $2.1\times10^{-2}$ & $8.2\times10^{-2}$  \\
      
      % U-NO (2022) & 0.0258 & 0.1713 & 0.0113 & 0.0078 & 0.0100 \\
      % F-FNO (2023) & 0.0263 & 0.2322 & 0.0077 & 0.0078 & 0.0070 \\
      % \textbf{LSM} & \textbf{0.0218} & \textbf{0.1535} & \textbf{0.0065} & \textbf{0.0059} & \textbf{0.0050} \\
      \midrule
      \textbf{Sp$^2$GNO (ours)} & $2.6\times10^{-2}$ & $\bm{1.6\times10^{-1}}$ &  $\bm{9.0\times10^{-3}}$ & $\bm{9.8\times10^{-3}}$  \\
      \bottomrule
    \end{tabular}
  }
\end{table}

\section{Conclusion}\label{sec:concl}

In this study, we presented the Spatio-Spectral Graph Neural Operator (Sp$^2$GNO), a cutting-edge neural architecture designed to address the challenges inherent in existing operator learning methods, particularly when applied to unstructured grids and irregular domains. By synergistically combining the localized processing capabilities of spatial graph neural networks with the global feature extraction strength of spectral graph neural networks, Sp$^2$GNO offers a robust solution that captures both local and global dependencies within complex domain graphs. This dual capability not only circumvents issues like over-smoothing and over-squashing typically associated with deep spatial GNNs but also mitigates the high computational costs linked with spectral GNNs, striking a balance between performance and efficiency.

The versatility of Sp$^2$GNO is evident from its performance in solving both time-dependent and time-independent partial differential equations across a variety of domain geometries, including those with irregular and complex structures. In rigorous benchmarking tests, Sp$^2$GNO consistently demonstrated superior accuracy, achieving the best results in three out of four test cases and ranking second in the remaining one. These results underscore its potential as a powerful tool for a wide range of real-world applications, particularly in computational mechanics and scientific computing.

One of the key technical decisions in this work was the use of the k-nearest neighbor (KNN) method for constructing the domain graph, with the value of 
$k$ serving as a critical hyperparameter. While this approach provided a solid foundation for the current implementation, we recognize that it may not always produce the most optimal graph topology for learning, which could, in turn, impact the overall performance of the model.
To further enhance the capabilities of Sp$^2$GNO, future research will focus on developing an adaptive extension that can learn the graph topology directly from the coordinate information of the domain geometry. This advancement aims to optimize the graph construction process, potentially leading to even greater accuracy and efficiency in solving partial differential equations across diverse and complex domains. By continuously refining and expanding the capabilities of Sp$^2$GNO, we aspire to contribute significantly to the field of scientific machine learning, providing more powerful and adaptable tools for tackling some of the most challenging problems in computational science.

\section*{Acknowledgements}
SC acknowledges the financial support received from the Ministry of Ports and Shipping via letter number ST-14011/74/MT (356529) and the Science and Engineering Research Board via grant number CRG/2023/007667.

\section*{Code availability}
Upon acceptance, all the source codes to reproduce the results in this study will be made available on request
%to the public on GitHub by the corresponding author.

\section*{Competing interests} 
The authors declare no competing interests.
% \clearpage
% \newpage
% \bibliographystyle{plain}
% % % You need to keep the file "references.bib" in the same folder as this ".tex". Citation in BibTex format are to be added in references.bib file. You may rename the file as you wish but you will have to change the following command accordingly then. 
% \bibliography{references.bib}
% % If you wish to add appendix to your research paper un-comment following three lines.

% \appendix
% \newpage

% \section{Appendix A: Additional Numerical Example}

\end{document}